\documentclass{article}
\usepackage{arxiv}
\usepackage{color, colortbl}
\usepackage{booktabs}
\usepackage{subcaption}
\usepackage{graphicx}
\usepackage{multirow}
\usepackage{multicol}
\usepackage{cite}
\usepackage{soulutf8}
\usepackage{url}
\usepackage{amsfonts}
\usepackage{amsmath}

\definecolor{lightgray}{rgb}{0.7,0.7,0.7}

\title{A Cluster-Matching-Based Method for Video Face Recognition}

\author{
  Paulo Renato C. Mendes\\
  Department of Informatics\\
  PUC-Rio\\
  Rio de Janeiro, Brazil\\
  \texttt{pmendes@inf.puc-rio.br} \\
  \And
 Antonio José G. Busson\\
 Department of Informatics\\
  PUC-Rio\\
  Rio de Janeiro, Brazil\\
  \texttt{busson@telemidia.puc-rio.br} \\
  \And
 Sérgio Colcher\\
 Department of Informatics\\
  PUC-Rio\\
  Rio de Janeiro, Brazil\\
  \texttt{colcher@inf.puc-rio.br} \\
  \And
 Daniel Schwabe\\
 Department of Informatics\\
  PUC-Rio\\
  Rio de Janeiro, Brazil\\
  \texttt{dschwabe@inf.puc-rio.br} \\
  \And
 Álan Lívio V. Guedes\\
 Department of Informatics\\
  PUC-Rio\\
  Rio de Janeiro, Brazil\\
  \texttt{alan@telemidia.puc-rio.br} \\
  \And
 Carlos Laufer\\
 Department of Informatics\\
  PUC-Rio\\
  Rio de Janeiro, Brazil\\
  \texttt{laufer@globo.com} \\
}

\begin{document}
\maketitle
\begin{abstract}
Face recognition systems are present in many modern solutions and  thousands of applications in our daily lives.
However, current solutions are not easily scalable, especially when it comes to the addition of new targeted people. 
We propose a cluster-matching-based approach for face recognition in video.
In our approach, we use unsupervised learning to cluster the faces present in both the dataset and targeted videos selected for face recognition.
Moreover, we design a cluster matching heuristic to associate clusters in both sets that is also capable of identifying when a face belongs to a non-registered person.
Our method has achieved a recall of 99.435\% and a precision of 99.131\% in the task of video face recognition.
Besides performing face recognition, it can also be used to determine the video segments where each person is present.

\end{abstract}


\maketitle

\section{Introduction}
\label{sec:intro}

In recent years, the popularity of platforms for the storage and transmission of video content has enabled a massive volume of video data production and consumption.
As an example, in 2019, more than one billion hours of YouTube videos were watched per day.\footnote{https://kinsta.com/blog/youtube-stats/}
Generating metadata with the identity information of the people present in a video can facilitate video indexing and retrieval.

Face Recognition has been an active research topic for many years~\cite{survey66, video2019webmedia}.
Many methodologies have been proposed, most commonly relying on comparing selected facial features of a given image with features of faces within a database.
Using only one sample reference image of a person's face for the comparison may result in classification errors due to factors related to variations in lighting, image resolution, angle, etc.~\cite{598229}.
To overcome this problem, some face recognition approaches use multiple face samples for comparison. However, this strategy does not scale well as the complexity is a function of the number of samples.
Other approaches treat the face recognition task as a classification problem~\cite{dadi2016improved, ghosal}, where a classifier model learns rules to assign faces to previously known classes within a dataset, where each class corresponds to one person.
Nonetheless, this kind of approach does not deal well when new classes are incorporated because of the need to retrain the classification models.
Moreover, when dealing with video, these kinds of methods have to be applied to each frame, again increasing the complexity.

In this work, we propose a cluster-matching-based approach for video face recognition where clustering is used to group faces in both the face dataset and in the target video.
Consequently, classes do not have to be previously known, and the effort spent with annotations is significantly reduced --- as it is done over clusters instead of single images.
Face recognition becomes a task of comparing clusters from the dataset with the ones extracted from images or video sources.
Therefore, our approach is easily scalable and can be used to automatically generate video metadata.

The remainder of this work is structured as follows. 
In Section \ref{sec:related_work}, we discuss some previous work that closely relates to ours.
In Section \ref{sec:dataset}, we describe the dataset we produced to perform experiments. 
The, Section \ref{sec:method}, we detail our proposed method, followed by two sections with experiments: 
Section \ref{sec:clustering_validation}, regarding the clustering methods we use, and 
Section \ref{sec:matching_validation}, with the experiments with our matching heuristic.
Section \ref{sec:face_video} is devoted to the overall evaluation of our method and, finally, 
in Section \ref{sec:final_remarks}, we conclude by discussing our results.

\section{Related Work}
\label{sec:related_work}

Face detection and recognition have been attracting the attention of researchers for more than two decades. Since the deep learning boom, face detection and recognition performance have greatly improved in terms of both speed and accuracy~\cite{masi2018deep}. Nowadays, face recognition systems are used for video surveillance and security systems, video analytics systems, smart shopping, automatic face tagging in photo collections, investigative tools that search for identities in social networks based on face images, and in thousands of other applications in our daily lives.

Traditional deep learning models for face recognition such as DeepFace~\cite{taigman2014deepface} and DeepID~\cite{sun2014deep} use a CNN with fully-connected layer output to produce a representation of high-level features (face embeddings) from an input image, followed by a softmax layer to indicate the identity of classes. Other approaches, such as FaceNet~\cite{schroff2015facenet}, can directly measure the similarity among faces using euclidean space. Inspired by DeepID, this model uses the \textit{triplet loss} as the loss function to estimate similarity to one character's face to a  collection of other faces. Triplet loss improves the accuracy of the  CNN output by minimizing the euclidean distance between the anchor and the positive (face of the same identity) while maximizing the distance between the anchor and the negative (face of another identity). In this work, we evaluated different pre-trained CNN backbones on VGGFace2 dataset~\cite{cao2018vggface2} to generate the face embeddings. This model is the state-of-the-art\footnote{https://paperswithcode.com/paper/vggface2-a-dataset-for-recognising-faces} in the face verification task on the IJB-B dataset~\cite{whitelam2017iarpa}.

Proprietary systems for face recognition and matching are widely used by social network platforms. For instance, Facer~\cite{hazelwood2018applied} is Facebook's face detection and recognition framework. Given a photograph, it first detects all the faces. Then, it runs a  deep model to determine the likelihood of that face belonging to one of the top-N user friends. This allows  Facebook to suggest which friends the user might want to tag within the uploaded photographs. FindFace\footnote{https://findface.br.aptoide.com/app} is an app that matches photos to profile pictures on VKontakte,\footnote{https://vk.com/} a Russian social networking website similar to Facebook. FindFace uses a deep model developed by NTech Lab that won the \textit{2017 IARPA Face Recognition Prize Challenge} (FRPC)~\cite{grother20172017}  in two nominations out of three (“Identification Speed” and “Verification Accuracy”). Similarly, our method can detect faces in videos and automatically recognize their identities by a clustering-based algorithm that uses a knowledge base with the faces pre-identified as a reference; however, a comparison with such methods was not possible due to access restrictions.

Some recent work is focused on video face recognition. Pena \textit{et al.}~\cite{globofacestream} proposed a face recognition system to detect characters within videos, called~\textit{Globo Face Stream}. Their method uses a Histogram of Oriented Gradients (HOG) feature combined with a linear classifier to detect faces. Next, they use  FaceNet to generate the embeddings, followed by the euclidean distance calculus to measure the similarity among faces. Rolim and Porto ~\cite{webmedia1} proposed a face-identification system that can be used in Virtual Learning Environment (VLE) systems to identify students through a webcam. Its structure is based on a client-server configuration that combines stages of processing facial recognition modules with a mechanism that monitors remote users. Their approach for facial recognition uses a technique based on feature selection of DCT coefficients and KNN for classification. 

Yang \textit{et al.}~\cite{yang2017neural} proposed a deep network for video face recognition called NAN (Neural Aggregation Network). They use a CNN to generate the embeddings, followed by an aggregation module that consists of two attention blocks which adaptively aggregate the feature vectors to form a single feature inside the convex hull spanned by them. Rao \textit{et al.}~\cite{rao2017attention} proposed a method for video face recognition based on attention-aware deep reinforcement learning. They formulated the process of finding the attention of videos as a Markov decision process and training the attention model without using extra labels. Unlike existing attention models, their method takes information from both the image space and the feature space as the input to make use of face information that is discarded in the feature learning process. Sohn \textit{et al.}~\cite{sohn2017unsupervised} proposed an adaptative deep learning framework for image-based face recognition and video-based face recognition. Given an embedding generated by a CNN, their framework adaptation is achieved by (1) distilling knowledge from the network to a video adaptation network through feature matching, (2) performing feature restoration through synthetic data augmentation, and (3) learning a domain-invariant feature through an adversarial domain discriminator. 

Like~\cite{globofacestream, yang2017neural, rao2017attention, sohn2017unsupervised}, our method uses a CNN to generate face embeddings from face images, with the difference that it uses an unsupervised cluster-based method to compare the similarity among face datasets and faces extracted from videos. Also, our approach can detect faces that do not have an identity registered in the face dataset with excellent performance.

\section{Dataset}
\label{sec:dataset}
Our dataset was collected using the information provided by the Brazilian Chamber of Deputies\footnote{\url{https://www.camara.leg.br/}} for the 55th legislature, which was in session from February 1st, 2015 through January 31st, 2019.
In total, 623 different deputies participated during some period in the 55th legislature, but we collected only 513 deputies from this set.

For each of the 513 deputies, we collected images in which he/she was present using Google images.
The resulting dataset has a total of 9,003 images, with a mean of ~17.55 images per deputy and a standard deviation of ~6.91. 
The maximum and minimum number of images per deputy are respectively 32 and 2.

We have randomly selected one image of each deputy for validation and the rest for training.
Thus, our dataset is divided into two: the training set, with 8,490 images, and the validation set, with 513 images.  

Besides, we created another set containing images of people who are not present in the deputies set.
For that, we randomly selected 513 images from the \emph{Labeled Faces in the Wild}~(LFW)~\cite{LFWTech} dataset, that contains more than 13,000 images of faces collected from the web,
and defined this subset of LFW as our \emph{Non-registered people} set.

\section{Method}
\label{sec:method}

Our method intends to recognize people in video using CNNs and clustering algorithms.
For didactic purposes we decided to divide our exposition in two phases: (i)~\emph{labeled clusters generation phase} and (ii)~\emph{cluster matching for video face recognition}, which are described in Sections~\ref{subsec:labeled_clusters} and \ref{subsec:cluster_match} respectively.

\subsection{Labeled Clusters Generation}
\label{subsec:labeled_clusters}

\begin{figure*}[!ht]
    \centering
    \includegraphics[width=\textwidth]{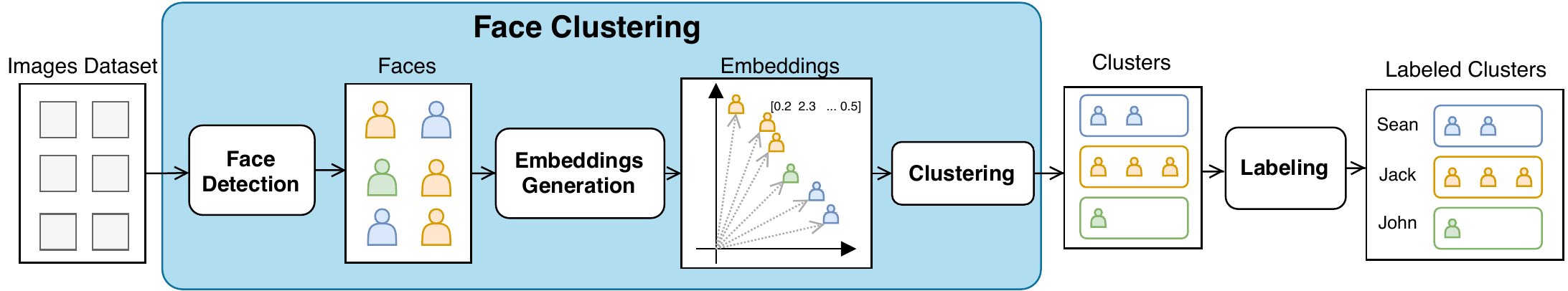}

    \caption{Labeled clusters process.}

    \label{fig:labeled_clusters}
\end{figure*}

The objective of this phase is to generate a set of labeled clusters. 
It is divided into four steps: \emph{face detection}, \emph{embeddings generation}, \emph{clustering} and \emph{labeling}. 
Figure \ref{fig:labeled_clusters} shows the pipeline we propose for this phase, described in the remainder of this subsection.

The \textit{Face Detection} step uses an object detection model for detecting faces in each of its images.
In general, any object detection model can identify which, among a known set of objects, are present in the image, and provides information about their positions.
In our case, objects are faces and, therefore, the face detection model is responsible for returning the bounding boxes of the faces present in the image, specified by the $x$ and $y$ axes coordinates of the upper-left corner and of the lower-right corner of the rectangle that establishes the visual limits that encapsulate each face. 
With these bounding boxes, we can isolate and extract the bounded images, obtaining a dataset composed of images of faces only.

The objective of the \textit{Embeddings Generation} step is to represent each face image as a vector space in $\mathbb{R}^{n}$.
To achieve that, it processes each of the faces generated in the previous step through a CNN, generating embeddings. 
An embedding is a representation of the input in a lower dimensionality space.
Ideally, an embedding captures some semantics of the input, e.g. by placing semantically similar inputs close together in the embedding space.
At the end of this step, we have all faces represented as embeddings.

In the \textit{Clustering} step, we group embeddings and, consequently, faces that are close in the embedding space using a clustering algorithm. 
Clustering is the task of dividing a set of data points, embeddings in this case, into a number of groups~(clusters) such that data points in a given group are similar to other data points in the same group and dissimilar to the data points in other groups.
The clustering process should produce a partition of the dataset, i.e., each generated cluster represents a specific person, and the union of all clusters covers the whole dataset.

Finally, in the \textit{Labeling} step, we assign labels~(identities) to represent the clusters.
A label can be anything that represents the faces present in the cluster, e.g. a name or an id number. 
Using this pipeline, instead of having to label every single face for constructing a labeled dataset, it is only necessary to label each generated cluster.
Consequently, all the faces present in a cluster are assigned to the same label. 
Hence, the complexity of labeling becomes a function dependent on the number of clusters, which is at most as great as the number of individuals.
At the end of this step, we have a dataset of labeled clusters.

\subsection{Cluster Matching for Video Face Recognition}
\label{subsec:cluster_match}

\begin{figure*}[!ht]
    \centering
    \includegraphics[width=\textwidth]{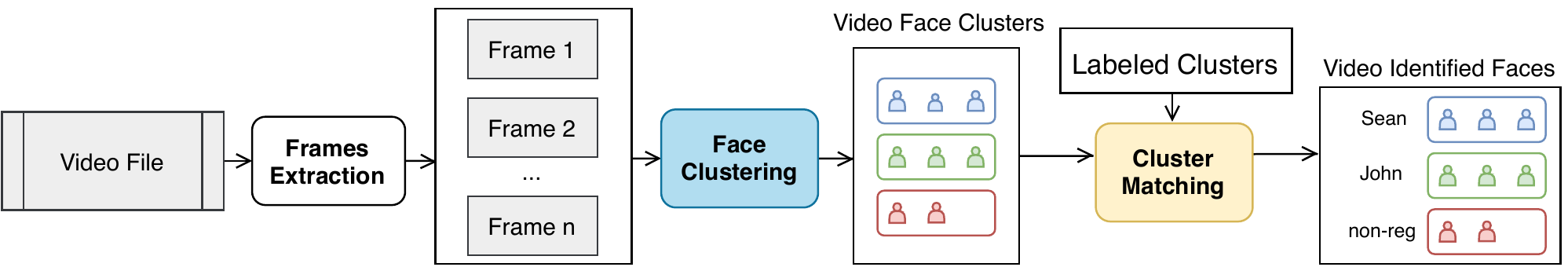}

    \caption{Cluster-matching based method for automatic face recognition in video files.}

    \label{fig:cluster_matching}
\end{figure*}

This phase aims at recognizing the faces present in a video file. 
It is divided in three steps: \emph{frames extraction}, \emph{face clustering} and \emph{cluster-matching}.
Figure \ref{fig:cluster_matching} shows the pipeline we propose for this phase, described in the remainder of this subsection.

First, we perform \textit{Frames Extraction} by receiving a video file as input and extracting its frames according to a given frame rate. 
These frames are used as a set of images for the next step.

Next, the \textit{Face Clustering} step, which is a macro-step that comprises the three first steps of the \emph{Labeled Clusters Generation} phase~(\emph{Face Detection}, \emph{Embeddings Generation}, and \emph{Clustering}), receives this set of images and returns a set of clusters of the faces present in the images received  (see Figure \ref{fig:labeled_clusters}).
%%
%In this phase, the input images are the frames of the video being processed.
%%
It is important to notice that this phase uses the same methods used for \emph{Labeled Clusters Generation}.
Consequently, the embeddings of the faces from the video are part of the same embedding space as the data points (faces) in the labeled clusters.

Finally, the \textit{Cluster Matching} step receives the set of clusters from the video and the set of labeled clusters, which is used as a reference for recognizing the clusters (and consequently the faces) in the video.
We designed a method based on cluster distance for performing this recognition.
We compare each candidate cluster in the video with each of the labeled clusters in the reference dataset, using what we call a \emph{cluster embedding}. This \emph{cluster embedding} is computed for both the candidate cluster in the video and the reference cluster in the labeled dataset.
The \emph{cluster embedding} maps each cluster to a vector in the embedding space of the face embeddings.
In the experiments, we evaluate two ways of obtaining a \emph{cluster embedding}.

Let $q$ denote the \emph{cluster embedding} of the query cluster, where $q~\in~\mathbb{R}^{n}$, in which $n$ is the dimension of the embedding space.
Let $K$ denote the set of labeled clusters. For each $k~\in~K$, let $c_k$ denote the \emph{cluster embedding} of $k$. 
We compute a similarity function of $q$ and each $c_k$ for $k~\in~K$. 
This similarity function is based on the Root Mean Square Error~(RMSE) function, which is largely used for computing embeddings distances in machine learning techniques.\footnote{https://www.sciencedirect.com/topics/engineering/root-mean-square-error}
We decided to use the inverse of the RMSE since we want to return larger values for clusters that are closer to $q$, to be able to use these values to compute a probabilistic distribution:

\begin{equation}
\label{equation:similarity_raw}
    s_{q,k} = \frac{1}{\sqrt{\frac{1}{n}\sum_{i=0}^{n}{(q_i-c_{k,i})^2}}}
\end{equation}

Since we are not interested in clusters that too distant from $q$, we define $\overline{s}_{q,k}$ that considers only the $\alpha$ largest $s_q$ values. The $\alpha$ value is a parameter that we further evaluate in this work. 
Thus, our similarity function becomes

\begin{equation}
\label{equation:similarity}
\overline{s}_{q,k}=\begin{cases}s_{q,k} & \text{if}~s_{q,k}~\in max_{\alpha(s_q)}\\0 & \text{otherwise}\end{cases}
\end{equation}

Given the similarity~$\overline{s}_{q,k}$, we compute the probability of $q$ being a match with each labeled cluster using the \emph{softmax} function.
This function takes as input a vector of real numbers and normalizes it into a probability distribution consisting of probabilities proportional to the exponential of the input numbers. 
It is defined as

\begin{equation}
\label{equation:probability}
    p_{q,k} = \frac{e^{\overline{s}_{q,k}}}{\sum_{j~\in~K}{e^{\overline{s}_{q,j}}}}
\end{equation}

Finally, we define the function $\sigma$ that, given a cluster embedding query $q$, returns the cluster in the labeled clusters whose \emph{cluster embedding} is more likely to have a match with the query if it has a probability greater than $0.5$. Otherwise, the query is assigned as being a match with none of the clusters ($\text{\o}$ is given as result). The $\sigma$ function is defined as follows.

\begin{equation}
\label{equation:sigma}
    \sigma{(q)} = \begin{cases}argmax(p_q) & if~~~p_{q,argmax(p_q)}~>~0.5\\\text{\o} & otherwise\end{cases}
\end{equation}

The $argmax$ function applied over $p_q$ returns the cluster $k$ whose probability of $q$ being a match with it is the greatest. 
We observed that when a cluster has faces that belong to a person present in the labeled clusters, the probability tends to be higher for one single labeled cluster.
However, when the person is not in any of the labeled clusters, the probability tends to be more distributed among different labeled clusters.
For that reason, when none of the labeled clusters has a probability greater than $0.5$, we say that the query cluster does not match any of the labeled clusters.
Hence, the person in the video represented by the query cluster is not present in the labeled clusters.
\section{Faces Clustering Validation}
\label{sec:clustering_validation}

In this section, we evaluate the quality of the clustering generated in the macro-step \emph{Faces Clustering}.
This step is used in both the \emph{Labeled Clusters Generation} phase~(Figure \ref{fig:labeled_clusters}) and \emph{Cluster Matching for Video Face Recognition} phase~(Figure \ref{fig:cluster_matching}).

In the \emph{Face Detection} step, we use MTCNN~\cite{mtcnn} (Multitask
Cascaded Convolutional Networks) which is widely used for the face detection task~\cite{mtcnn1, mtcnn3}.
We used the  \emph{mtcnn} Python library for its implementation.\footnote{\url{https://pypi.org/project/mtcnn/}}

We tested three different CNNs for the \emph{Embeddings Generation} step in association with three different clustering algorithms for the \emph{Clustering} step.
For that, we evaluated each pair of \emph{CNN x ClusteringAlg} through the quality of the clusters they generated. 
To experiment, we used the training set with 8490 deputy images.

%In what follows, we describe the algorithms used for the experiments, the metrics used to evaluate and the results we obtained.

\subsection{Embeddings Generation}
\label{subsec:exp_setup}

For this step, we have three candidate CNNs that were previously trained on the VGGFace2 dataset~\cite{cao2018vggface2}. 
The VGGFace2 dataset contains $3.31$ million images of $9131$ subjects and has large variations in pose, age, illumination, ethnicity, and profession.
The three candidate CNNs used are VGG-16~\cite{vgg16}, ResNet-50~\cite{resnet} and SE-ResNet-50~\cite{senet}~(SeNet-50 for short). VGG-16 generates embeddings in the $\mathbb{R}^{512}$  feature space, while ResNet-50 and SeNet-50 generate embeddings in the $\mathbb{R}^{2048}$ feature space.

%We used the architecture and weights pre-trained on the VGGFace2 dataset\cite{cao2018vggface2} available in the \emph{keras-vggface} library  \footnote{\url{https://github.com/rcmalli/keras-vggface}}. 

\subsection{Clustering Algorithms}
\label{subsec:clustering_algs}
For this step, we selected the following clustering algorithms as candidates: k-Means~\cite{lloyd1982least}, affinity propagation~\cite{frey2007clustering}, and agglomerative clustering~\cite{ward1963hierarchical}. Each of these algorithms is briefly explained next.

K-Means~\cite{lloyd1982least} is one of the most widely used unsupervised machine learning algorithms. 
To process the data to be clustered, the K-Means algorithm begins with a randomly selected group of $K$ centroids, which are updated iteratively to optimize the distance of the data points to the closest centroid.
The algorithm stops when either the centroids have stabilized or the maximum number of iterations has been reached.

The Affinity Propagation algorithm~\cite{frey2007clustering}, in contrast with other clustering methods, does not require the number of clusters to be previously specified.
In this algorithm, each data point sends messages to the other points informing them of their relative attractiveness to the sender. 
Those targets then reply to the senders informing their availability to associate with them, considering the attractiveness of all the messages it received.
The senders then reply to the targets informing the target's updated relative attractiveness.
This process continues until a consensus is established.
When the sender data point is associated with one of its targets, that target becomes the point's exemplar.
The points with the same exemplar are assigned to the same cluster.

The Agglomerative Clustering algorithm recursively merges the pair of clusters that minimally increase a given linkage distance.
The linkage criteria specify the distance to use between two sets of data points.
The Agglomerative Clustering algorithm~\cite{ward1963hierarchical} merges pairs of clusters that minimize the linkage criteria.
In this work, we chose the \emph{Ward} criteria~\cite{ward1963hierarchical}, which minimizes the variance of the clusters being merged.
By using this method, at each step, the algorithm finds the pair of clusters that leads to a minimum increase in total within-cluster variance after merging.
This increase is measured by a weighted squared distance between cluster centers.
In the first step, each data point is a cluster.
The clusters are merged following the criteria until the number of clusters $K$ is reached.

Different from the Affinity Propagation algorithm, K-Means and Agglomerative Clustering require the number of clusters in advance. For this case, we used 513~(number of deputies in the \emph{train set}).
However, when the number of cluster is not known, a strategy for defining it is required for these two algorithms.

\subsubsection{Strategy for unknown number of clusters}
\hspace{.1pt}
\label{subsec:unknown_nclusters}

In this subsection, we describe a strategy based on the \emph{Silhouette Score} that is used when the number of clusters~(number of people) is not known in advance. 

The \emph{Silhouette Score}~\cite{rousseeuw1987silhouettes} corresponds to the mean of the \emph{Silhouette Coefficient} of all samples.
This coefficient ($S$) for each sample is 
\begin{equation}
\label{equation:Silh}
    S = \frac{b-a}{max(a,b)} 
\end{equation}
where $a$ is the mean distance from a sample to all other samples in the same cluster, and $b$ is the mean distance from a sample to all other samples in the closest cluster to that sample.
In this way, the best value is 1 and the worst is -1. Values close to 0 indicate overlapping clusters, whereas negative values usually indicate that a sample has been assigned to the wrong cluster since a different cluster is more similar.

In this strategy, we try to increase the number of clusters until the maximum Silhouette Score does not increase for more than $t$ times in a row or until it reaches the maximum number of clusters, which is the number of data points.
When it stops, we return the clustering configuration with the highest Silhouette Score.
Since the Silhouette Coefficient requires at least two clusters, it would not be possible to compute the Silhouette Score for a clustering configuration with only one cluster~(there are only faces of a single person).
To overcome this problem, we start with 2 clusters consecutively increasing it as described above. Then, if the returned clustering configuration has a Silhouette Score smaller than $0.2$, that provably indicates overlapping, we say that all faces belong to one single cluster.

\subsection{Metrics}\label{sec:metrics}

We evaluate the models using the V-Measure~\cite{vmeasure}, which is an entropy-based measure that computes how successfully the criteria of homogeneity and completeness have been satisfied. This metric is extensively used for comparing clustering solutions and has been used in different domain fields such as biology~\cite{bio1}, computational linguistics~\cite{nlp1}, and document engineering~\cite{doceng}.

The V-Measure is a harmonic mean of homogeneity and completeness scores, similar to how precision and recall are frequently combined into F-measure~\cite{van1979information}. It assumes a dataset comprising $N$ data points, and a set of classes, $C = \{c_i|i = 1,..., n\}$ and a set of clusters, $K = \{k_i|i = 1,...,m\}$. They also assume $A$ as a matrix produced by the clustering algorithm representing the clustering solution, such that $A = \{a_{ij}\}$ where $a_{ij}$ is the number of data points that are members of class $c_i$ and elements of cluster $k_j$.

The homogeneity is perfect when a clustering algorithm assigns only those data points that are members of a single class to a single cluster, so that the entropy is zero in each cluster.

In this way, the homogeneity score determines how close a given clustering is to this ideal by examining the conditional entropy of the class distribution given the proposed clustering. 
Therefore, when the clustering is perfectly homogeneous, such a value, $H(C|K) = 0$. 
 
On the other hand, when the clustering is not perfect according to this criterion, this value is proportional to the size of the dataset and the classes.
For this reason, the authors normalize this value by the maximum reduction in entropy the clustering information could provide, specifically, $H(C)$.
$H(C|K)$ has its maximal value when it is equal to $H(C)$ and provides no new information.

Finally, to address to the convention of $1$ being desirable and $0$ undesirable, the authors define homogeneity as:

\begin{equation}
\label{equation:homo}
    h = \begin{cases} 1 & \text{if}\ H(C|K) = 0 \\1-\frac{H(C|K)}{H(C)} & \text{otherwise}\end{cases}
\end{equation}

where

\begin{equation}
    H(C|K) =  -\sum_{k=1}^{|K|}{\sum_{c=1}^{|C|}{\frac{a_{ck}}{N}log\frac{a_{ck}}{\sum_{c=1}^{|C|}{a_{ck}}}}}
\end{equation}

\begin{equation}
    H(C) = -\sum_{c=1}^{|C|}{  \frac{\sum_{k=1}^{|K|}a_{ck}}{n}log \frac{\sum_{k=1}^{|K|}a_{ck}}{n} }
\end{equation}

Completeness is symmetrical with respect to homogeneity, and it is perfect when a clustering assigns all data points that are members of the same class to a single cluster. 
For computing such a score, the authors examine the distribution of cluster assignments within each class.
Completeness is defined as 

\begin{equation}
\label{equation:completeness}
    c = \begin{cases} 1 & \text{if}\ H(K|C) = 0 \\1-\frac{H(K|C)}{H(K)} & \text{otherwise}\end{cases}
\end{equation}

where

\begin{equation}
    H(K|C) =  -\sum_{c=1}^{|C|}{\sum_{k=1}^{|K|}{\frac{a_{ck}}{N}log\frac{a_{ck}}{\sum_{k=1}^{|K|}{a_{ck}}}}}
\end{equation}

\begin{equation}
    H(K) = -\sum_{k=1}^{|K|}{  \frac{\sum_{c=1}^{|C|}a_{ck}}{n}log \frac{\sum_{c=1}^{|C|}a_{ck}}{n} }
\end{equation}

Finally, by using Equation \ref{equation:homo} and Equation \ref{equation:completeness}, V-measure is defined as 
\begin{equation}
    V_{\beta} = (1+\beta)\frac{h\cdot c}{\beta\cdot{h+c}}
\end{equation}

The parameter $\beta$ is used to calibrate the relative importance of homogeneity and completeness when computing the V-measure. If $\beta$ is greater than 1, the completeness is more important for the V-measure, whereas 
if $\beta$ is less than 1, homogeneity is more important for the V-measure. 
In this work, we use $\beta = 1$, so that completeness and homogeneity contribute equally to the V-measure.

\subsection{Results}
\label{subsec:results}

%\hl{In this subsection}, we show the results obtained with the embeddings generated by each of the \emph{CNNs} in association with each of the clustering algorithms. 
%%
Table \ref{tab:results_clustering} shows the homogeneity, completeness, and V-Measure for each combination of CNN and clustering algorithm.

\begin{table}[!ht]
\centering
\small
\caption{Results of the evaluation of the clusters created by each combination of CNN and clustering algorithms.}
\begin{tabular}{@{}ccccc@{}}

\toprule
\textbf{CNN} & \textbf{Clustering} & \textbf{$h$} & \textbf{$c$} & \textbf{$V_1$} \\ \midrule
                  & KM                  & 0.9665                     & 0.9675                      & 0.9670             \\
ResNet-50         & AP                  & 0.0000                     & 1.0000                      & 0.0000             \\
                  & AC                  & 0.9821                     & 0.9798                      & 0.9810             \\ \midrule
                  & KM                  & 0.9725                     & 0.9726                      & 0.9725             \\
SeNet-50          & AP                  & 0.9859                     & 0.9558                      & 0.9706             \\
                  & \textbf{AC}         & \textbf{0.9862}            & \textbf{0.9833}             & \textbf{0.9847}    \\ \midrule
                  & KM                  & 0.8340                     & 0.8415                      & 0.8378             \\
VGG-16            & AP                  & 0.0000                     & 1.0000                      & 0.0000             \\
                  & AC                  & 0.8899                     & 0.8929                      & 0.8914             \\
\end{tabular}
\label{tab:results_clustering}

\end{table}
From Table \ref{tab:results_clustering}, one can conclude that the best combination of CNN and clustering algorithm was SeNet-50 with Agglomerative Clustering.
For this reason, we decided to use SeNet-50 for the \emph{Embeddings Generation} phase and the Agglomerative Clustering algorithm for the \emph{Clustering} phase.
As a result, each face on the dataset is represented as an embedding in the $\mathbb{R}^{2048}$ produced by SeNet-50.

One can observe that the Affinity Propagation algorithm had a V-measure of $0$ when used with ResNet-50 and VGG-16. 
This happens because, in both cases, the algorithm assigned all the data points~(face embeddings in this context) to one single cluster. 
Consequently, those combinations had a homogeneity score of $0$ because there was no information gain after clustering--and completeness of 1 because data points of the same class were not scattered in different clusters.
With such a combination, the V-measure, which is a harmonic mean of the homogeneity and completeness, has a value of $0$.

\section{Cluster-Matching Validation}
\label{sec:matching_validation}

Based on the results of the experiment described in Section \ref{sec:clustering_validation}, we developed another experiment in order to choose the best $\alpha$ value in Equation \ref{equation:similarity} and the best method for computing the \emph{cluster embedding} for the clusters.

As the best combination for clustering was SeNet-50 with the Agglomerative Clustering algorithm, this experiment uses the embeddings and labeled clusters generated by it.
Hence, the labeled clusters are the set of faces from the 8490 deputy images with their respective embeddings and labels.
Each cluster was labeled with the name of the deputy whose face is more frequent in it.

For testing, we used the \emph{Validation set}~($V$) consisting of 513 deputy images and the \emph{Non-registered people set}~($U$) also consisting of 513 images.
We used these images as if they were the clusters from a video file, each cluster having a single sample.
In this way, we have a more heterogeneous set for performing this experiment as if we had used a small set of video files.

%In what follows we describe the methods for computing \emph{cluster embeddings} used for the experiment, the metrics used to evaluate and the results we obtained.

%%\subsection{Cluster embedding selection}

We evaluate two methods for computing a \emph{cluster embedding} for the labeled clusters: \emph{cluster centroid} and \emph{sample with best silhouette coefficient}. A cluster centroid denotes the mean of the elements in a cluster.
It is calculated the mean of all elements in a cluster for each axis in the embedding space. The second method uses the embedding of the sample whose silhouette coefficient is the highest.
This coefficient is calculated using the mean intra-cluster distance and the mean nearest-cluster distance for each sample~(detailed in Subsection \ref{subsec:unknown_nclusters}).

\subsection{Metrics}

We evaluate the two methods for computing the \emph{cluster embedding} by assessing how well the cluster matching recognizes registered people and correctly tells when a person is not registered.
A non-registered person should not have a match with any of the labeled clusters.

Let $\lambda(q)$ be the most frequent label of a cluster $q$, $|V|$ the size of the \emph{Validation set} and $|U|$ the size of the \emph{Non-registered people set}.
Although $\lambda$ represents the label of a cluster in general, we also define $\Lambda$ as being the set of all the face labels present in a cluster.
Let us also assume true as $1$ and false as $0$.
To perform the evaluation, we compute the following four metrics:

\noindent\begin{minipage}[c]{0.45\linewidth}
    \begin{equation}
    m_1 = \sum_{v~\in~V}{\frac{\sigma(v)\ne\text{\o}}{|V|}}
    \end{equation}
\end{minipage}
\hfill
\begin{minipage}[c]{0.45\linewidth}
    \begin{equation}
    m_2 = \sum_{v~\in~V}{\frac{\lambda(v) = \lambda(\sigma(v))}{|V|}}    
    \end{equation}
\end{minipage}

\noindent\begin{minipage}[c]{0.45\linewidth}
    \begin{equation}
    m_3 = \sum_{v~\in~V}{\frac{\lambda(v)~\in~\Lambda(\sigma(v))}{|V|}} 
    \end{equation}
\end{minipage}
\hfill
\begin{minipage}[c]{0.45\linewidth}
    \begin{equation}
    m_4 = \sum_{u~\in~U}{\frac{\sigma(u)=\text{\o}}{|U|}}
    \end{equation}
\end{minipage}

Where $m_1$ denotes the percentage of clusters from $V$ matched with any cluster, $m2$ the percentage of clusters from $V$ matched with a cluster with the same label, $m3$ the percentage of clusters from $V$ matched with a cluster in which the label is present, and $m4$ that denotes the percentage of clusters from $U$ that have not matched with any cluster.

\subsection{Results}

In this subsection we show the results for each of the methods for generating the \emph{cluster embeddings} in the \emph{cluster matching} step.
For each of the methods, we tested different $\alpha$ values in Equation \ref{equation:similarity}.

\begin{table}[!ht]
\centering
\small
\caption{Results obtained using cluster centroids as \emph{cluster embeddings}}

\label{tab:results_centroid}
\begin{tabular}{ccccc}

%\hline

\toprule
\textbf{$\alpha$} & \textbf{m1} & \textbf{m2} & \textbf{m3} & \textbf{m4} \\ \hline
2 & 100.000\% & 94.932\% & 98.246\% & 0.000\% \\
3 & 97.076\% & 94.152\% & 96.881\% & 94.542\% \\
4 & 95.517\% & 93.177\% & 95.322\% & 98.635\% \\
\textbf{5} & \textbf{94.542}\% & \textbf{92.398}\% & \textbf{94.347}\% & \textbf{99.220}\% \\
6 & 93.567\% & 91.813\% & 93.372\% & 99.610\% \\
7 & 92.788\% & 91.228\% & 92.593\% & 99.805\% \\
8 & 91.813\% & 90.643\% & 91.618\% & 99.805\% \\
9 & 90.448\% & 89.279\% & 90.253\% & 99.805\% \\
10 & 89.474\% & 88.304\% & 89.279\% & 99.805\%
\end{tabular}

\end{table}

Table \ref{tab:results_centroid} shows the results obtained using the cluster centroids as \emph{cluster embeddings}.
It can be seen that with the higher the $\alpha$ value is, the lower is the percentage of registered faces assigned to the correct labeled cluster.
On the other hand, the percentage of non-registered faces assigned to none of the labeled clusters increases with the $\alpha$ value.
One can see that with $\alpha = 5$, we have a balance between assigning people to the correct cluster and being able to tell when a person is not in the labeled clusters.

\begin{table}[!ht]
\small

\caption{Results obtained using samples with the highest silhouette coefficient as \emph{cluster embeddings}}

\label{tab:results_silhouette}
\centering
\begin{tabular}{ccccc}
\toprule
\textbf{$\alpha$} & \textbf{m1} & \textbf{m2} & \textbf{m3} & \textbf{m4} \\ \hline
2 & 100.000\% & 93.177\% & 94.347\% & 0.000\% \\
3 & 86.355\% & 85.575\% & 86.160\% & 98.246\% \\
4 & 76.998\% & 76.608\% & 76.998\% & 99.415\% \\
5 & 72.904\% & 72.515\% & 72.904\% & 99.805\% \\
6 & 68.811\% & 68.421\% & 68.811\% & 100.000\% \\
7 & 64.912\% & 64.522\% & 64.912\% & 100.000\% \\
8 & 61.209\% & 60.819\% & 61.209\% & 100.000\% \\
9 & 60.039\% & 59.649\% & 60.039\% & 100.000\% \\
10 & 58.285\% & 57.895\% & 58.285\% & 100.000\%
\end{tabular}

\end{table}

Table \ref{tab:results_silhouette} shows the results obtained using the sample with the highest silhouette coefficient as the \emph{cluster embedding} for each of the labeled clusters.
One can observe that the $\alpha$ value correlates with the percentage values similar to the one observed in Table \ref{tab:results_centroid}.
By using the silhouette coefficient for choosing the \emph{cluster centroids}, we have an algorithm more capable of determining when a person is not registered, achieving a percentage of $100\%$ when $\alpha>5$.
However, when using this approach, the algorithm fails to correctly assign registered people to the correct labeled cluster in comparison to when it uses the cluster centroids.

\section{Video Face Recognition Evaluation}
\label{sec:face_video}

To evaluate our complete pipeline, we selected videos that contain only registered people (videos \emph{a} to \emph{d}), videos with both registered and non-registered people (videos \emph{e} to \emph{i}) and videos with only non-registered people (videos \emph{j} to \emph{m}).
The labeled clusters are the same used in Section \ref{sec:matching_validation}, which were generated using MTCNN~\cite{mtcnn} for detecting faces, SeNet-50 for extracting its embeddings and the Agglomerative Clustering algorithm to cluster them. 
The label of the clusters corresponds to the name of the deputy whose face is more frequent in it.

For performing the face identification on a video file, we first extract its frames using a frame rate of 1fps. 
For each of the frames, we extract the faces present on it using MTCNN~\cite{mtcnn}. 
Then, for each face identified, we extract its embedding using SeNet-50 and cluster these faces using the Agglomerative Clustering algorithm.
Since we do not know the number of people present in the video, we use the strategy described in Subsection \ref{subsec:unknown_nclusters} with the maximum time of sequential increases $t=5$, that was empirically determined to be a good stop point.

Next, we perform the \emph{Cluster Matching} with the labeled clusters and the video face clusters.
We used cluster centroids as \emph{cluster embeddings} and a $\alpha=5$ as the results of Section \ref{sec:matching_validation} show that this configuration is able to correctly match the clusters while preserving the capacity of distinguishing non-registered faces.
At the end of this process, each face present on the video is labeled either with the name of a registered person or as non-registered.

We evaluate our method by the Precision~(Prec), Recall (Rec), and F1-Score for the faces in the video. 
As usual, the Precision is defined as the percentage of detected faces that our method correctly labels, 
the Recall gives the percentage of faces that our method correctly labels among all faces in the video, and 
the F1-score represents an overall performance metric based on the  harmonic mean of the precision and recall.
\begin{table}[!ht]
%\small
\caption{Results using the proposed approach in videos.}

\label{tab:results_videos}
\centering
\begin{tabular}{@{}clccccccc@{}}
\toprule
\textbf{Video} & \textbf{URL} & \textbf{\#P} & \textbf{\#R} & \textbf{\#F} & \textbf{\#EM} & \textbf{Rec.} & \textbf{Prec.} & \textbf{F1} \\ \rowcolor{lightgray}
\multicolumn{9}{c}{videos with only registered people}\\
a & \url{https://youtu.be/QjTZ_TE1U_g}
& 1 & 1  & 105 & 105 & 100.000\% & 100.000\% & 100.000\%  \\
b & \url{https://youtu.be/4D5GGR3g_7c} 
& 1 & 1  & 80  & 79  & 100.000\% & 98.765\%  & 99.379\%   \\
c & \url{https://youtu.be/quYNTUsOTb8}
& 1 & 1 & 60  & 59   & 100.000\% &	98.113\% & 99.048\%   \\
d & \url{https://youtu.be/eB6kJYaoxHc}
& 2 & 2  & 226 & 226 & 100.000\% & 100.000\% & 100.000\%  \\ 
\rowcolor{lightgray}
\multicolumn{9}{c}{videos with both registered and non-registered people}\\
e & \url{https://youtu.be/j07yExfJ4mA}
& 2 & 1  & 101 & 99  & 100.000\% & 98.113\%  & 99.048\%   \\
f & \url{https://youtu.be/Db2I1uUyDlE}
& 2 & 1  & 650 & 650 & 100.000\% & 100.000\% & 100.000\%  \\
g & \url{https://youtu.be/sf56sWeiMyo}
& 8 & 1  & 201 & 190 & 96.471\%  & 96.850\%  & 96.660\%   \\
h & \url{http://youtu.be/dYAFXogdqW4}
& 4 & 1  & 231 & 215 & 98.097\%  & 98.514\%  & 98.305\%   \\
i & \url{http://youtu.be/NHglWWOKmc4}
& 2 & 1  & 88  & 88  & 100.000\% & 100.000\% & 100.000\%  \\ 
\rowcolor{lightgray}
\multicolumn{9}{c}{videos with only non-registered people}\\
j & \url{https://youtu.be/UH0nTHb6OdY}
& 6 & 0  & 625 & 617 & 99.551\%  & 99.551\%  & 99.551\%   \\
k & \url{http://youtu.be/wHN5vYlJ-Vk}
& 2 & 0  & 231 & 228 & 100.000\% & 98.872\%  & 99.433\%   \\
l & \url{http://youtu.be/WwRdjf4eEgk}
& 2 & 0  & 131 & 121 & 99.482\%  & 95.522\%  & 97.462\%   \\
m & \url{http://youtu.be/3dIVdsiPDH8}
& 1 & 0  & 225 & 222 & 99.556\%  & 99.115\%  & 99.335\%   \\
\bottomrule
\end{tabular}

\end{table}
We also count the number of exact frames match (\#EM). It corresponds to the number of frames (\#F) for which our method correctly labeled all the faces that appears. 
Table \ref{tab:results_videos} shows the results we obtained, the number of different people in each video~(\emph{\#P}), and the number of people who are present in the labeled clusters~(\emph{\#R}).

\begin{figure}[!ht]
\centering
    \begin{subfigure}{0.47\linewidth}
        \centering
        \includegraphics[width=0.9\textwidth]{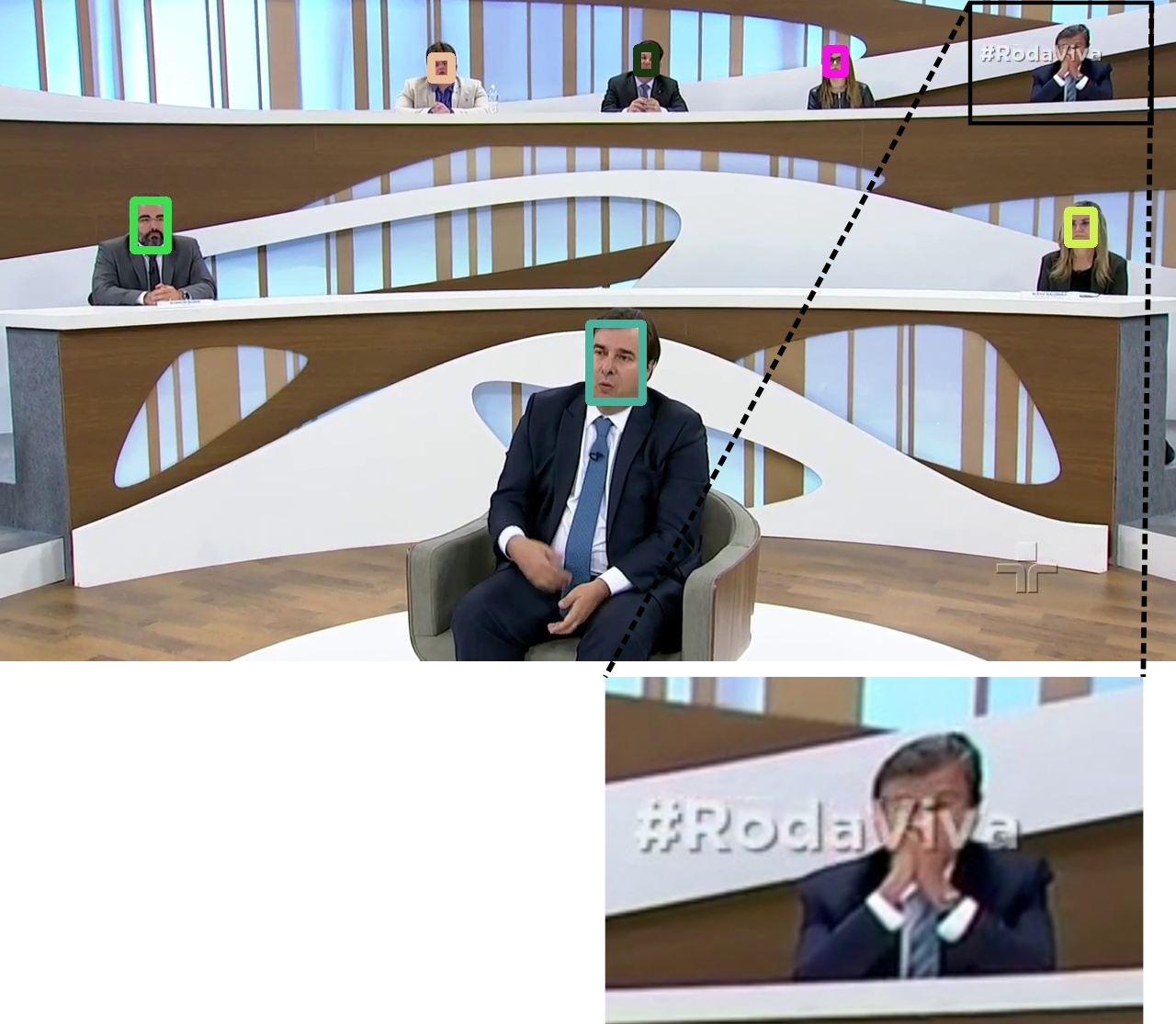}
        \caption{Example where a face is not detected because it is covered}
        \label{fig:recall}
    \end{subfigure}\hfill
    \begin{subfigure}{0.47\linewidth}
        \centering
        \includegraphics[width=0.9\textwidth]{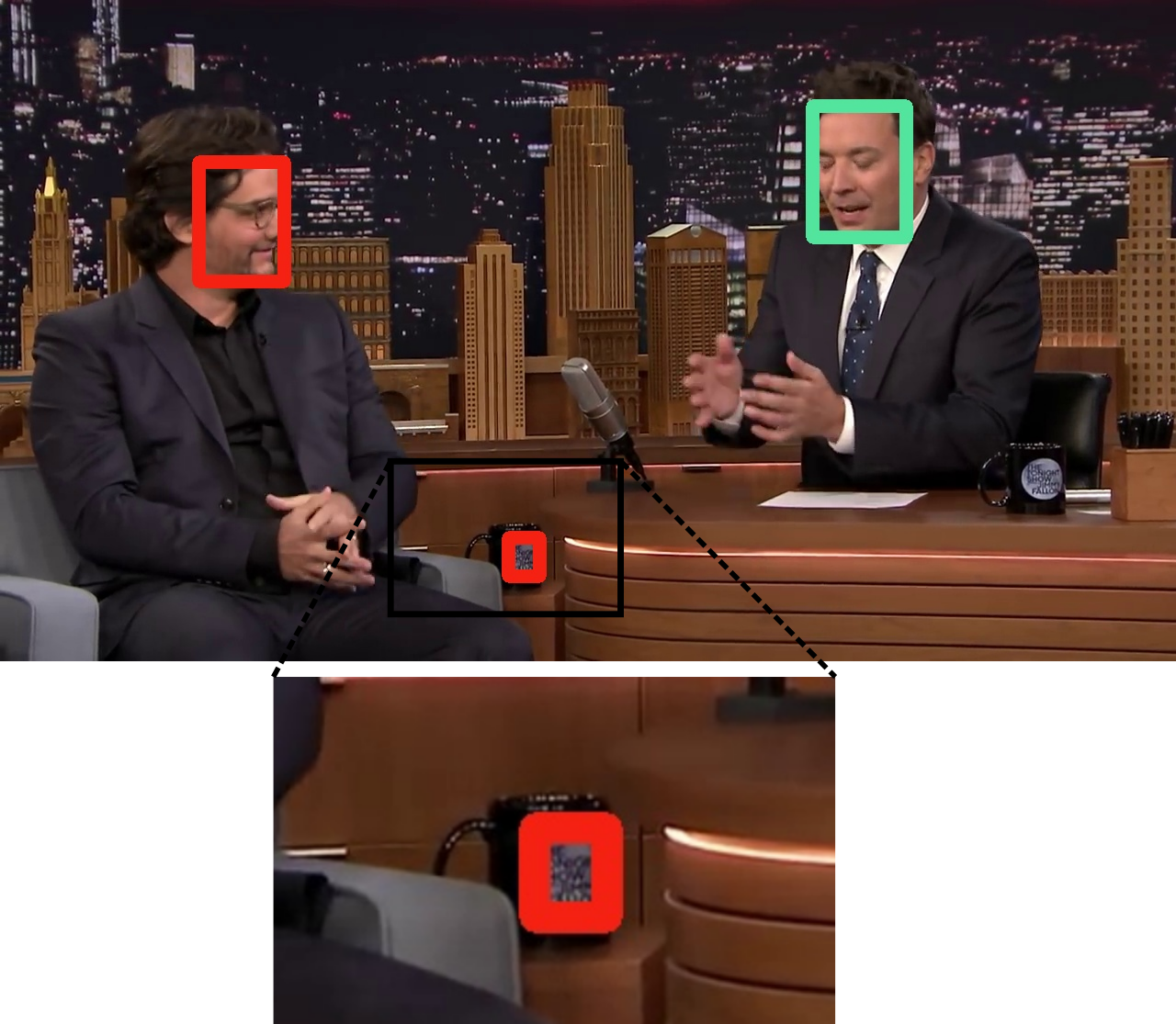}
        \caption{Example where a mug is detected as if it was a face.}
        \label{fig:precision}
    \end{subfigure}

\caption{Cases where our approach was incorrect.}

\end{figure}

One can observe that our method is able to correctly classify the faces of people that are present in the labeled clusters while also being able to tell when a person is not registered in the labeled clusters.
However, Table \ref{tab:results_videos} shows that the precision was smaller for some videos. Analyzing these videos, we observed that this result was mostly due to some non-face objects that were detected as if they were faces.
In \emph{Video l}, for instance, in some frames, a mug was detected as if it was a face~(Figure \ref{fig:precision}). 
On the other hand, in \emph{Video g}, that had the lowest recall, one face was not detected in some frames because it was partially covered by a text~(Figure \ref{fig:recall}).

\begin{figure}[!ht]
    \centering
    \includegraphics[width=0.8\linewidth]{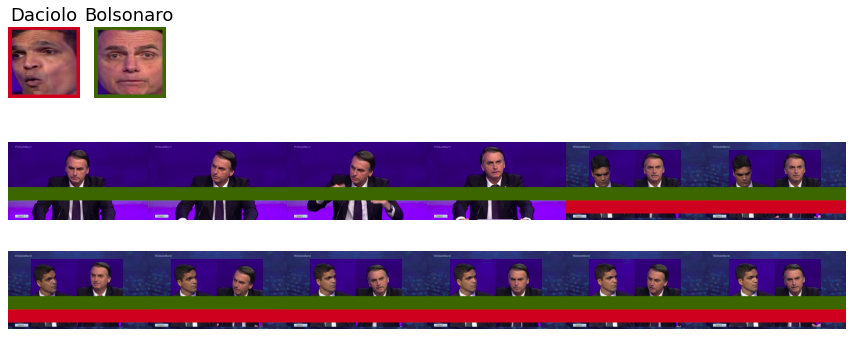}

    \caption{Timeline with tagged frames by their clusters of registered people}
    \label{fig:timeline_pol}

\end{figure}

Similar to the work of Pena \emph{et. al}~\cite{globofacestream}, our method can be used to generate metadata in video files indicating the people that appear in it.
Figure \ref{fig:timeline_pol} shows the first 12 seconds of \emph{Video d}~(detailed in Table \ref{tab:results_videos}) where two identified Brazilian politicians are shown in each frame with its respective colored cluster.

\begin{figure}[!ht]
    \centering
    \includegraphics[width=0.8\linewidth]{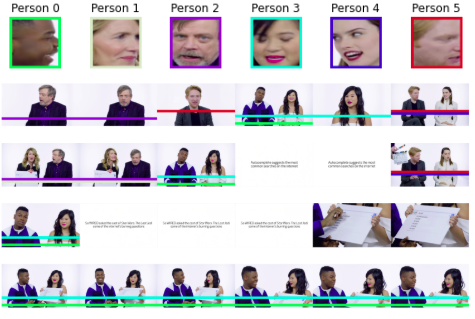}

    \caption{Timeline with tagged frames by their clusters of non-registered people}

    \label{fig:timeline}
\end{figure}

Besides being able to recognize people in video files, by using face embeddings and clustering, we can detect the frames where the same person appears without even knowing who the person is or if he/she is in the labeled clusters.
This can be done by following the pipeline described in Figure \ref{fig:cluster_matching} up to the \emph{Face Clustering} step, obtaining the video face clusters.
Figure \ref{fig:timeline} shows the first 24 seconds of \emph{Video j}~(detailed in Table \ref{tab:results_videos}) with the frames tagged with the clusters identified in each frame, where each color represents a cluster.
\section{Final Remarks}
\label{sec:final_remarks}
This work proposed a cluster-matching-based method for video face recognition. 
This method uses face embeddings, clustering algorithms, and a heuristic for cluster matching in order to recognize people in video. 
Our method has achieved a recall of 99.435\% and a precision of 99.131\% when considering all faces present in the frames extracted from a set of 13 video files.  
Moreover, our method is also capable of determining the video segments where each person is present.
With this information, it is capable of generating metadata regarding the temporal presence of actors in a video.

As a consequence of face clustering, our work also demonstrates that this technique can be useful for labeling datasets in a less time-consuming way, where clusters are labeled instead of individual data points.
Although this work focuses in faces, the method proposed can be applied to other domain fields.
The only requirement is that the designer has to  conceive a technique in which objects of the same class are placed closed to each other in the corresponding vector space.

One of the limitations of this work is related to the size of the video set used for the overall evaluation of our method. 
This is due to the difficulty of finding videos where it is possible to manually identify in each frame whether each person present is registered or not in our labeled clusters.
Other limitation is that the proposed method was not able to detect partially covered faces. 
This is due to a limitation of the MTCNN~\cite{mtcnn}, which is currently not robust enough to deal with this condition.
In future work, we intend to address these problems.

%% Inserir agradecimento caso aprovado

%\section*{ACKNOWLEDGEMENTS}
%This work was supported by the Coordination for the Improvement of Higher Education Personnel (CAPES), Brazil.

\bibliographystyle{hacm}
\bibliography{sample-base}

\end{document}